# Pre-processing training data improves accuracy and generalisability of convolutional neural network based landscape semantic segmentation

Andrew Clark[1,2,*], Stuart Phinn[1], Peter Scarth[1]

[1]Remote Sensing Research Centre, The University of Queensland, St Lucia, Brisbane Queensland, Australia

[2]Applied Agricultural Remote Sensing Centre, The University of New England, Armidale, New South Wales, Australia

[*]Corresponding author: Andrew.Clark@uqconnect.edu.au

## Abstract

In this paper, we trialled different methods of data preparation for Convolutional Neural Network (CNN) training and semantic segmentation of land use land cover (LULC) features within aerial photography over the Wet Tropics and Atherton Tablelands, Queensland, Australia. This was conducted through trialling and ranking various training patch selection sampling strategies, patch and batch sizes and data augmentations and scaling. We also compared model accuracy through producing the LULC classification using a single pass of a grid of patches and averaging multiple grid passes and three rotated version of each patch. Our results showed: a stratified random sampling approach for producing training patches improved the accuracy of classes with a smaller area while having minimal effect on larger classes; a smaller number of larger patches compared to a larger number of smaller patches improves model accuracy; applying data augmentations and scaling are imperative in creating a generalised model able to accurately classify LULC features in imagery from a different date and sensor; and producing the output classification by averaging multiple grids of patches and three rotated versions of each patch produced and more accurate and aesthetic result. Combining the findings from the trials, we fully trained five models on the 2018 training image and applied the model to the 2015 test image with the output LULC classifications achieving an average kappa of 0.84 user accuracy of 0.81 and producer accuracy of 0.87. This study has demonstrated the importance of data pre-processing for developing a generalised deep-learning model for LULC classification which can be applied to a different date and sensor. Future research using CNN and earth observation data should implement the findings of this study to increase LULC model accuracy and transferability. However, there are several outstanding elements which need to be address in future research including: evaluating models with additional LULC classes and training models for each class separately; altering the multiple pass strategy for producing the output classification to be more efficient through testing a reduced number of augments for each patch; and the integration of additional ancillary data such as climate and elevation data to give each pixel more context which will be important for studies over broader areas.

# 1 Introduction

## 1.1 Land Use and Land Cover Mapping

Remote sensing as a young science has already undergone several paradigm shifts, for example, from a plain pixel-based analysis to subpixel analysis and geographic object-based image analysis (Blaschke et al., 2014). Lately, the term 'big data', labelled as the fourth paradigm in science (Hey, 2009), has been used to describe challenges associated with data intensive sciences. The steadily increasing volume of data, such as remotely sensed multispectral imagery, leads to general big data problems where methods are required to process and analyse the input data for efficient, generalised, transferable, and accurate information extraction (Liu, 2015; Ma et al., 2015).

Advances in earth observation technologies have provided efficient and cost-effective land use and land cover (LULC) mapping, encompassing a larger area at a higher accuracy compared to traditional field surveys (Bai et al., 2017). As a result, many programs around the world were established based on manual digitisation in a Geographic Information System (GIS) environment using image interpretation with assistance from other ancillary data (Lillesand et al., 2015).

Currently, there are operational LULC mapping programs in several countries. These include the Global Land Cover Characteristics Database from the United States Geological Survey, Co-ORdinated INformation on the Environment (CORINE) from the European Environmental Agency, GeoBase from the Canadian Council on Geomatics and Natural Resources and the Australian Land Use and Management Program. These programs involve extensive manual interpretation of imagery and other ancillary data to derive LULC although there has been some automation of land cover classes.

## 1.2 Automated Land Use and Land Cover Classifications

Image classification is fundamental in LULC analysis since the early days of the remote sensing discipline (Jensen, 2007). There have been many studies exploring classification techniques for LULC classification; however, it is still unclear which are the best classifiers (Pandey et al., 2021).

Traditional approaches, such as maximum likelihood, fuzzy logic and object-oriented classifications, are referred to as shallow learning. These methods extract data based on spatial, spectral, textural, morphological and other cues (Ball et al., 2017). However, shallow learning analytical techniques for extracting LULC information using high spatial resolution imagery but of low spectral and temporal resolution, cannot successfully separate land use classes due to similar spectral signatures between features (Pandey et al., 2021).

In contrast, because deep-learning is multi-layered and learns from the data itself, results can be significantly more accurate than shallow learning (Deng, 2014; Ma et al., 2019) and it has been shown to outperform manual human editing (Zhang et al., 2016).

## 1.3 Deep-learning

Deep-learning for image segmentation has recently become the superior classification technique for earth observation data including identification of LULC in very high resolution (<1 m) data. This popularity is evident from the number of review papers attempting to bring order and clarity to the plethora of recent studies (Hoeser & Kuenzer, 2020; Kattenborn et al., 2021; Maxwell et al., 2021; Zang et al., 2021).

Convolutional Neural Networks (CNN), a form of deep-learning, can utilise contextual information as well as spectral information to undertake image analysis. CNN are the current network architectures of choice with U-Net (Ronneberger et al., 2015) being one of the most popular (Hoeser et al., 2020). The U-Net has been used in multiple LULC studies including forest coverage (Flood et al., 2019), urban studies such as building, road and car identification (Neupane et al., 2019) and agricultural applications (Clark & McKechnie, 2020).

Numerous studies have shown deep-learning techniques can successfully classify land use features; however, there are limited real world mapping applications as most literature surveyed was constrained geographically or restricted to a standard set of training images such as University of California Merced Land Use Dataset. Furthermore, there are few studies which focus on generalisation and assess accuracy when applying the model to data from another time, sensor, or geographical area (Maxwell et al., 2021).

### 1.3.1 Challenges for Deep-Learning Applications and Project Aims

There are several challenges new projects must overcome to use deep-learning techniques for automated data extraction from earth observations. Obtaining and processing training data is a major hurdle although there is an abundance of imagery (Burke et al., 2021). There are freely available existing datasets which can form the basis for training data such as those discussed in Section 1.1. However, these products are produced at a continental scale with resolutions greater than 30 metres and do not match the sub-metre spatial resolution imagery generally used with CNNs. Although some effort has been made recently to release higher resolution datasets (e.g., [https://github.com/chrieke/awesome-satellite-imagery-datasets](https://github.com/chrieke/awesome-satellite-imagery-datasets)).

In earth observation data and classifications, the disparity within and between classes causes class imbalances. Within a particular class, there can be a variety of factors such as environmental and climatic influences, solar angle and clouds and the influence of recent or absent rainfall, which need to be considered. Further to this, there can be variations within the class simply because of the elements from which it is composed. For example, the tree fruits class within the Australian Land Use and Management (ALUM) classification represents several different tree fruit crop types such as papaya, banana and mango which all have a different leaf and growth structure. Although increasing the classification resolution to the commodity level assists with class consistency, this can present other challenges as subtle variation between classes makes separation difficult using earth observation without additional information such as ground based observations. A balance is needed between what training data can be collected from satellite or aerial imagery and creating a robust model to account for variation within and between classes.

Another challenging factor within remote sensing applications is class representation of the landscape. Most applications tend to have one dominant class and several classes which make up only a small proportion of the landscape. Systematic or random generation of training patches will have very few training samples for underrepresented classes, resulting in poor classification for these. With one class dominating the image, it is highly likely that the classification algorithm will misclassify smaller classes without incurring a huge penalty. In addition, different areas within the training data class features results in features with small areas being less sampled than larger features, resulting in their poor classification.

To address LULC mapping in a consistent and repeatable way, this study aimed to establish standard training data processing recommendations which can be generally applied to high resolution RGB earth observation data prior to training a deep-learning model, including:

- how to sample the data,
- which patch size was the most effective,
- what effect the size of the batch of training data had on model training,
- how to ensure model transferability through data augmentations and scaling and
- how to create a more accurate and aesthetic classification through averaging the results of multiple prediction passes and augmentations.

Determining a standard set of pre-processing parameters for training data will assist future projects on how to approach deep-learning segmentation problems.

## 2 Methodology

### 2.1 Project Area

The project area, located in North Queensland, Australia, encompasses the towns of Mareeba in the north, Atherton in the south and Dimbulah in the west (Figure 1).

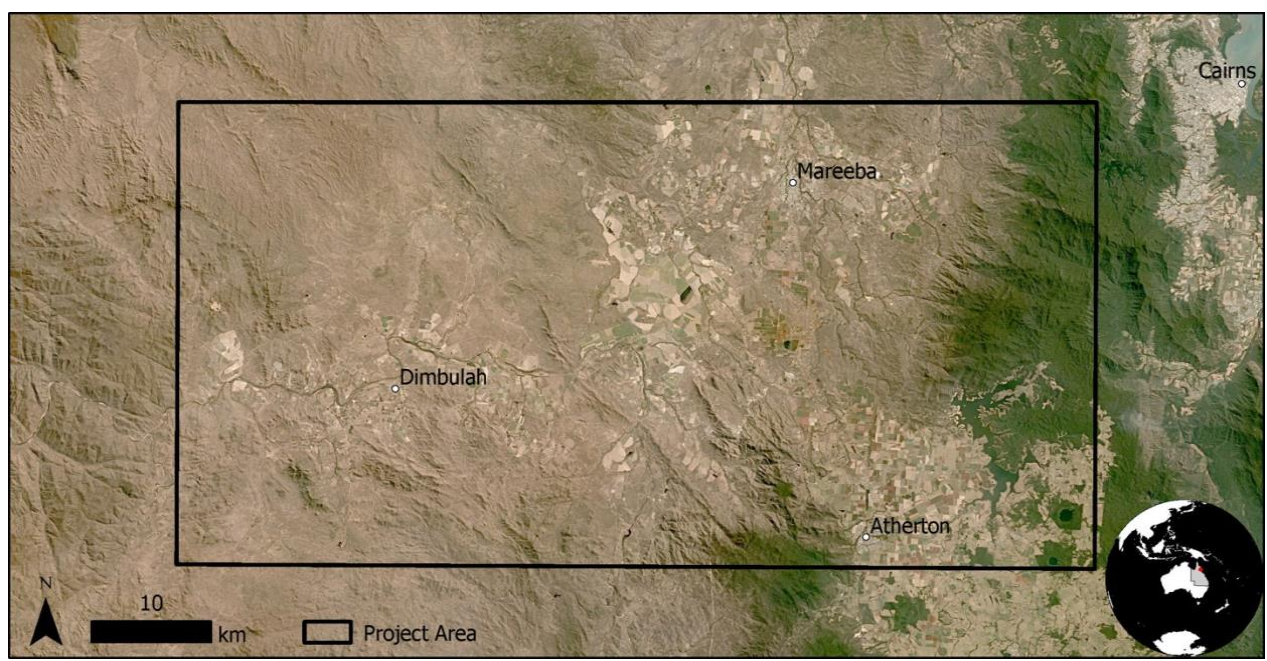

*Figure 1: Project area, North Queensland, Australia. Imagery basemap provided by the Queensland Government.*

The eastern part of the project area contains part of the world heritage listed Wet Tropics rainforests with the dominant land uses of production from relatively natural environments, conservation and natural environments, and production from irrigated agriculture and plantations (Table 1) (DSITI, 2017).

*Table 1: Project area primary land uses (DSITI, 2017).*

| Primary Land Use | Hectares | Proportion |
|---|---|---|
| Production from relatively natural environments | 201,625 | 67.3% |
| Conservation and natural environments | 40,404 | 13.5% |
| Production from irrigated agriculture and plantations | 37,939 | 12.7% |
| Intensive uses | 10,133 | 3.4% |
| Water | 7,219 | 2.4% |
| Production from dryland agriculture and plantations | 2,143 | 0.7% |
| **Total** | **299,463** | **100.0%** |

## 2.2 Image Data

Two orthorectified aerial imagery mosaics acquired under the Queensland Government Spatial Imagery Subscription Plan were used in the project (Figure 2). The mosaic used for training data and assessing each trial was acquired between 1st and 27th August 2018, referred to here as the 2018 training image. The training image was mostly captured using a Vexcel Ultracam Eagle camera except for the southwestern corner which was captured by and A3 Edge camera. The two cameras have different spectral properties which can be seen in Figure 2b. The test mosaic was acquired between 17th July and 14th October 2015 using an A3 Edge camera, referred to here as the 2015 test image.

The images were captured with a fixed-wing mounted three-band true-colour camera, at spatial resolutions of 25 cm and 20 cm for 2015 and 2018 respectively. The data were provided as an orthorectified mosaics. As shown in Figure 2, the quality of the imagery is not consistent across the study area. Unfortunately, the specific post-processing details were not listed within the supplied metadata, however, this was the highest resolution data available for the project area within the Queensland Government archive.

The data were resampled to 50 cm using cubic convolution to reduce data volume, reducing overall training time and to ensure the resolution was consistent between images.

## 2.3 Data collection

Eight LULC classes were selected along with a ninth 'Other' class covering all other land uses. The chosen classes fall within the *Production from irrigated agriculture and plantations* primary land use class presented in Table 1. Data were manually collected within the 2018 training image and 2015 test image by hand digitising polygons to best represent each feature's extent. The classes included banana plantations, berry crops, forestry plantations, sugarcane crops, mature tree crops, young tree crops, tea tree plantations and vineyards with all other remaining areas within the test area classified as other land uses. These classes were selected based on existing training data from previous work (Clark & McKechnie, 2020), prior knowledge of the area, ease of identification of the features within imagery and the ability to digitise to a high level of detail.

The 2018 data was used for training while the 2015 data was used as a comparison between the model and human classifications. An independent accuracy assessment is detailed in section 2.8.

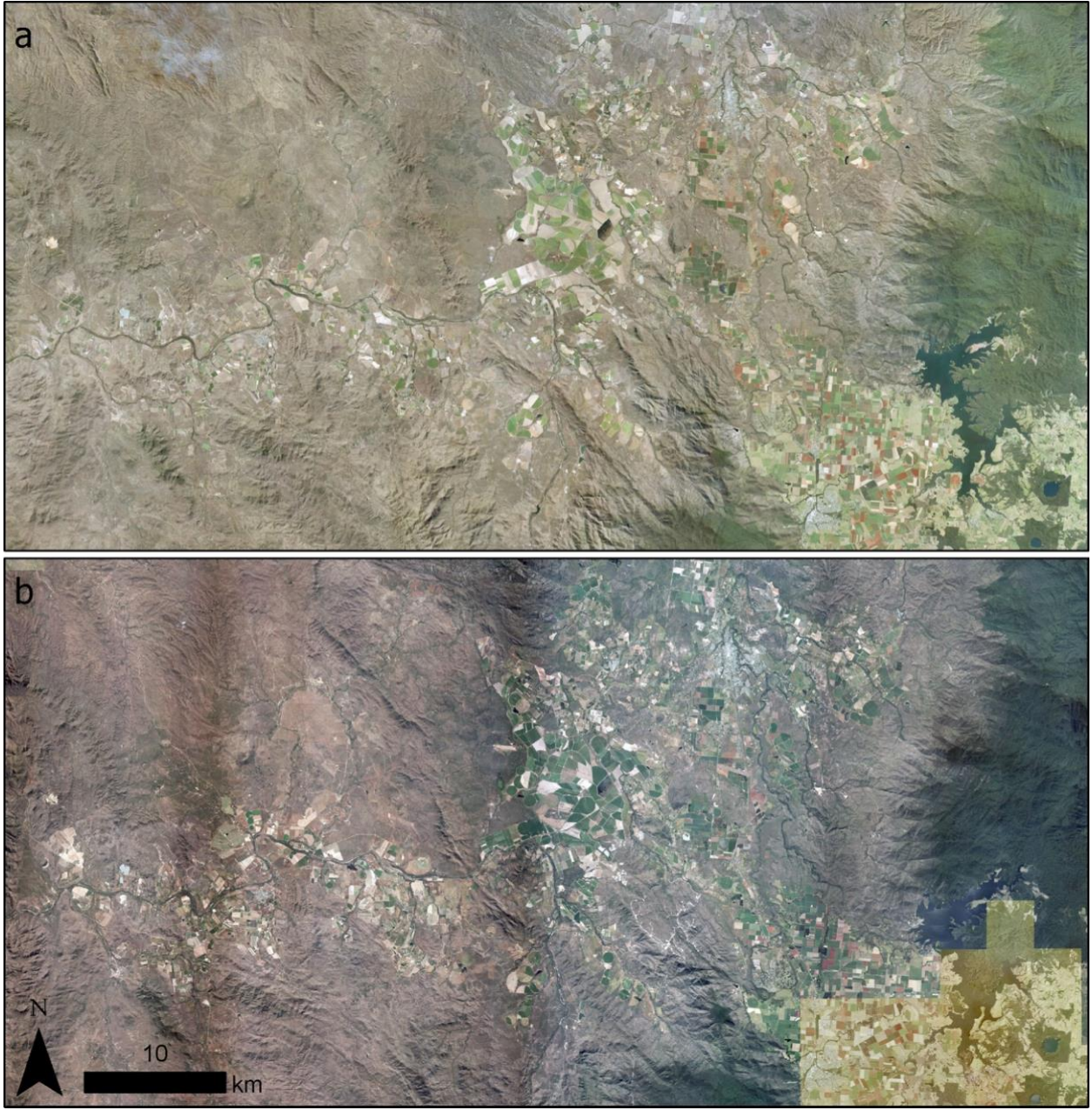

*Figure 2: Project area ortho-rectified aerial imagery for 2015 (a) and 2018 (b). Data supplied by the Queensland Government.*

## 2.4 Field verification

A field trip was conducted in January 2020 when 1,582 observations were made within the project area. These data were used in the verification and the refinement of the training data. However, due to time difference between the field observation and imagery acquisition date (17 months), these data were not used to assess the accuracy of the final classification.

The observations were collected using an Apple iPad Pro running Collector for ArcGIS. The data were contained within a feature service and stored within the ArcGIS Online servers. For each observation, the LULC type and growth stage was recorded. Optionally, additional fields allowed recording information such as land management, observation photo and other information within a comment field.

## 2.5 Deep-learning trials

The objective of this study was to determine a standard set of training processes which can be generally applied to earth observation data. These processes include training data sampling strategies, patch size and how to feed these data to the GPU (batch size, data scaling and augmentations) for learning (Figure 3). Additionally, we trialled derivation of the classifications based on the average of multiple inferences of the same test image. This was achieved through augmentation (image rotations) and by offsetting the start of the inference by half the number of pixels contained within a single patch.

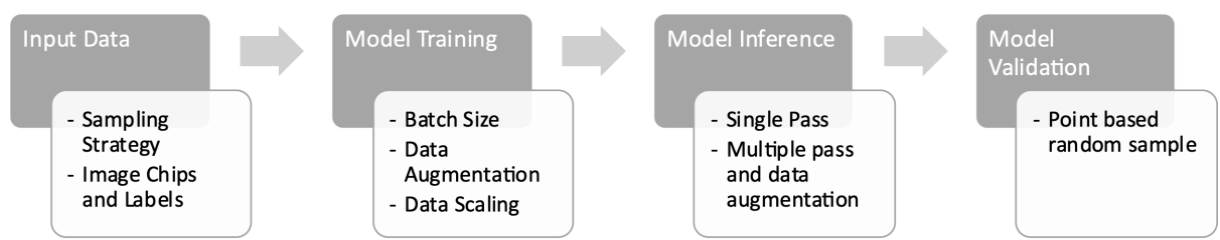

*Figure 3: Project trials, stages and validation steps.*

For this part of the study, all parameter trials were tested independently. Table 2 summarises the parameter trials, range of test values and default values when the parameter was not being tested. The only parameters altered within the trials were the ones to be tested. All other parameters remained consistent for the training. There was no default value for batch size as this was optimised according to and consistent within each trial parameter.

*Table 2: Parameter trial, number of patches, test values and defaults for the deep-learning trials.*

| Parameter | Number of training patches | Test Values | Default |
|---|---|---|---|
| Batch size | 3,986 | 10, 50, 100, 150, 200, 250, 280 | * |
| Patch size and sampling strategy | 2,408 – 154,514 | $128^2$, $256^2$, $512^2$, $1024^2$ | $512^2$ |
| Data Augmentations | 22,830 | True, False | False |
| Data Scaling | 22,830 | True, False | False |
| Multiple Pass Prediction | 22,830 | True, False | False |

Within each trial, multiple values of the parameters were tested by training five models for twenty epochs (or iterations of the training data) for each value. It was deemed that repeating the training a total of five times was sufficient to capture any random variance within the training process. Training the models for twenty epochs was not enough iterations of the data to produce a fully trained model; however, it was enough to give an indication of training performance and examine the training progress consistently over all tests while reducing the resource load.

The results can only be compared against each other within the same parameter trial and is not an indication of the accuracy of the classification from a fully trained model. Training each trial five times for twenty epochs produced an indication on how well a particular set of parameters performed against the 2018 training image; however, we refrained from assessing the accuracy against the 2015 test image until we had fully trained models based on the optimised parameters. The method to undertake fully training a model was discussed in section 2.10.

### 2.5.1 Batch Size

The batch size refers to the number of patches and labels which are processed on the GPU at one time and is limited by the size of the patches and GPU memory. Using a patch size of 512 x 512 pixels with three bands and corresponding nine band labelled data, the maximum number of patches which could be processed at one time was 280 so the trials for the batch size consisted of batches ranging from 10 to 280 (Table 2). As this trial involved training 35 models, the number of patches were limited to 3,986 to limit time and resource requirements.

### 2.5.2 Patch size and sampling strategy

Two training data sampling approaches were trialled: a systematic grid sampling strategy and a stratified random sampling approach based on area.

2.5.2.1 Grid sampling strategy

The grid sampling strategy is a systematic sampling approach. The image is divided into an even grid with each grid cell 1024 x 1024 pixels (512 m$^2$) in size. To reduce the number of patches, only grid cells which intersected the eight main training classes were retained. Patches which only intersected the 'other' class were excluded.

To test for optimum patch size, these grid cells formed the basis for subsequent grid cell sizes. Each 1024 x 1024 cell was divided into four cells to produce grid cell sizes of 512 x 512 pixels (256 m x 256 m). This process was repeated to produce a minimum patch size of 128 x 128 (64 m x 64 m). Figure 4 shows the distribution of the training data.

This approach ensures the same data were fed to the CNN but split into different sized patches. With every reduction in patch size, the number of patches increased four-fold.

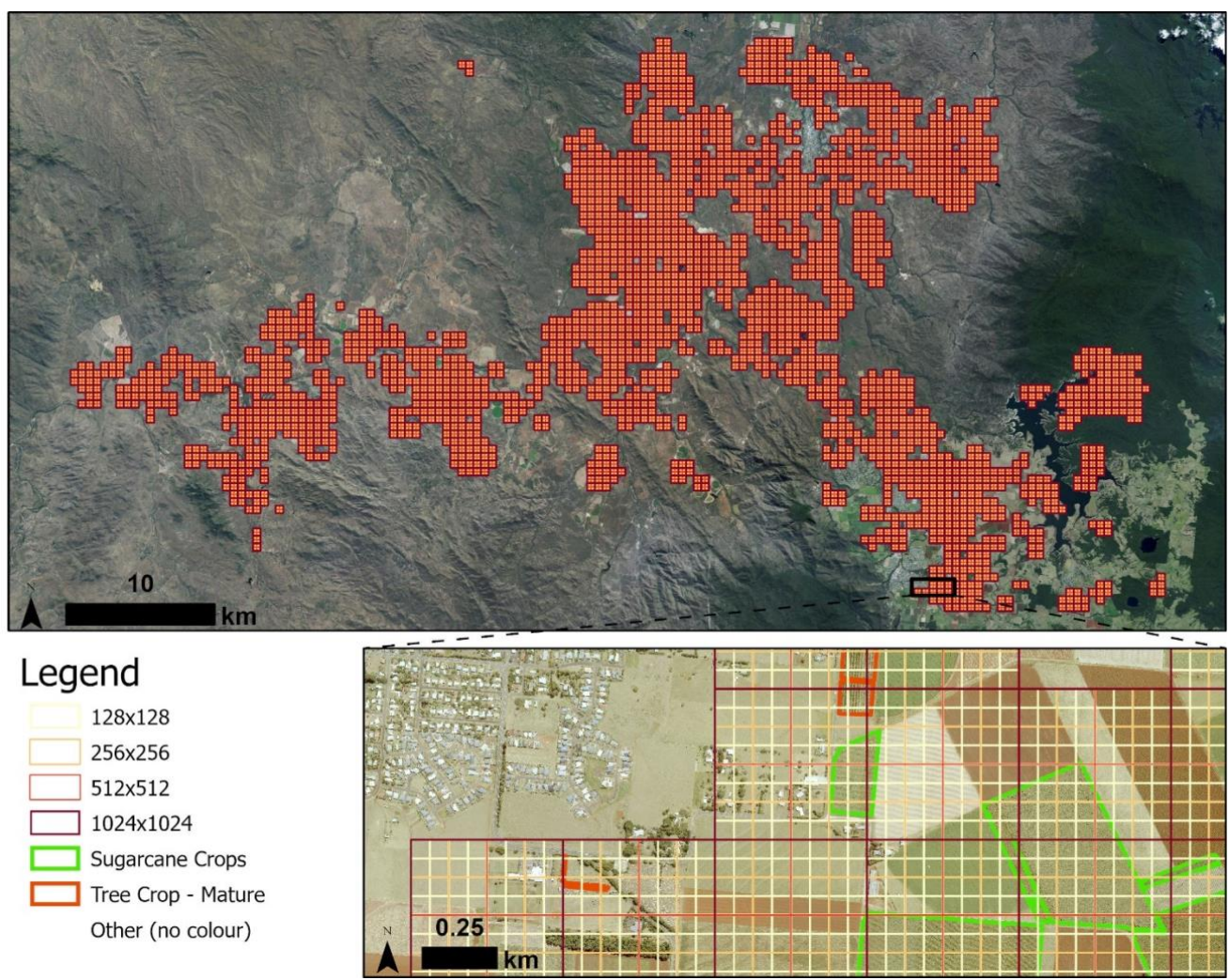

*Figure 4: Patch layout from the grid-based sampling strategy. Training classes show are derived from the manually derived training dataset.*

### 2.5.2.2 Stratified random sampling strategy

The systematic grid sampling strategy is a comprehensive sampling strategy; however, it does not account for the imbalance in class areas with the training data dominated by larger classes. To account for this disproportion as well as ensuring all features within a particular class are sampled at least once, a stratified random sampling approach based on area was developed.

For this sampling strategy, the number of patches for a particular class was calculated by multiplying the required number of patches (based on the result of the grid sampling strategy) by the log area of the target class and dividing by the sum of the log area for all classes. The result was rounded up to the nearest integer (Equation 1).

$$N_{cp} = \left\lceil N_p \frac{\log(a_c)}{\sum_n^i \log(a_c)} \right\rceil \quad \text{Equation 1}$$

where $N_{cp}$ is the number of class patches, $N_p$ is the total number of training patches, $a_c$ is the class area.

Each feature within the training data was sampled at least once to ensure all variations of the classes were captured. The number of patches generated within the feature was calculated by multiplying the number of class patches by the proportion of area that the feature represented of the total class rounded up to the nearest integer (Equation 2).

$$N_{fp} = \left\lceil N_{cp} \frac{a_f}{a_c} \right\rceil \quad \text{Equation 2}$$

Where $N_{fp}$ is the number of feature patches, $N_{cp}$ is the number of class patches, $a_f$ is the feature area and $a_c$ is the class area.

To generate the patch extent, a coordinate was randomly selected within the feature's geometry. The generated point formed the centroid of the patch.

To be consistent with the grid approach, a similar number of training patches ($N_p$) was produced. The stratified random sample approach rounded up the number of classes and feature patches and sampled every feature within the training data. As a result, the exact number of patches could not be matched without randomly excluding data which may disproportionately affect the classes. Table 3 lists the difference in patch number for each patch size.

Table 3: Number of patches for patch size and sampling strategy trials.

| Patch Size | Number of patches | | Difference | Difference (%) |
|---|---|---|---|---|
| | Stratified method | Grid method | | |
| $128^2$ | 154,514 | 154,112 | 402 | 0.26% |
| $256^2$ | 38,533 | 38,528 | 5 | 0.01% |
| $512^2$ | 9,635 | 9,632 | 3 | 0.03% |
| $1024^2$ | 2,412 | 2,408 | 4 | 0.17% |

Table 4 shows the number of patches for each class and patch size. The table highlights the differences between the methods and shows the number of patches in the larger classes (e.g., other and sugarcane crops) have decreased while the smaller classes (e.g., berry crops and vineyards) have increased.

Table 4: Number of patches per class for each patch size and sampling strategy trial.

| Class | $128^2$ | | $256^2$ | | $512^2$ | | $1024^2$ | |
|---|---|---|---|---|---|---|---|---|
| | grid | area | grid | area | grid | area | grid | area |
| Banana Plantations | 6,155 | 17,509 | 1,861 | 4,535 | 637 | 1,214 | 249 | 369 |
| Berry Crops | 387 | 14,050 | 129 | 3,567 | 49 | 892 | 21 | 230 |
| Other | 132,194 | 98,019 | 35,955 | 3,567 | 9,519 | 892 | 2,408 | 2,410 |
| Plantation Forestry | 3,604 | 16,565 | 1,100 | 4191 | 360 | 1050 | 133 | 266 |
| Sugarcane Crops | 23,705 | 19,585 | 6,994 | 5,162 | 2,247 | 1,503 | 795 | 473 |
| Tea Tree | 766 | 14,795 | 274 | 3,761 | 110 | 964 | 50 | 247 |
| Tree Crops - Mature | 26,360 | 24,152 | 9,005 | 7,151 | 3,414 | 2,531 | 1,418 | 941 |
| Tree Crops - Young | 4,361 | 17,304 | 1,634 | 4,636 | 722 | 1,423 | 367 | 451 |
| Vineyards | 546 | 14,539 | 194 | 3,693 | 84 | 933 | 43 | 241 |
| **Total** | **154,112** | **154,514** | **38,528** | **38,533** | **9,632** | **9,635** | **2,408** | **2,412** |

Figure 5 shows the spatial distribution of the stratified sampling method and illustrates the clustering of patches around classes with a smaller area.

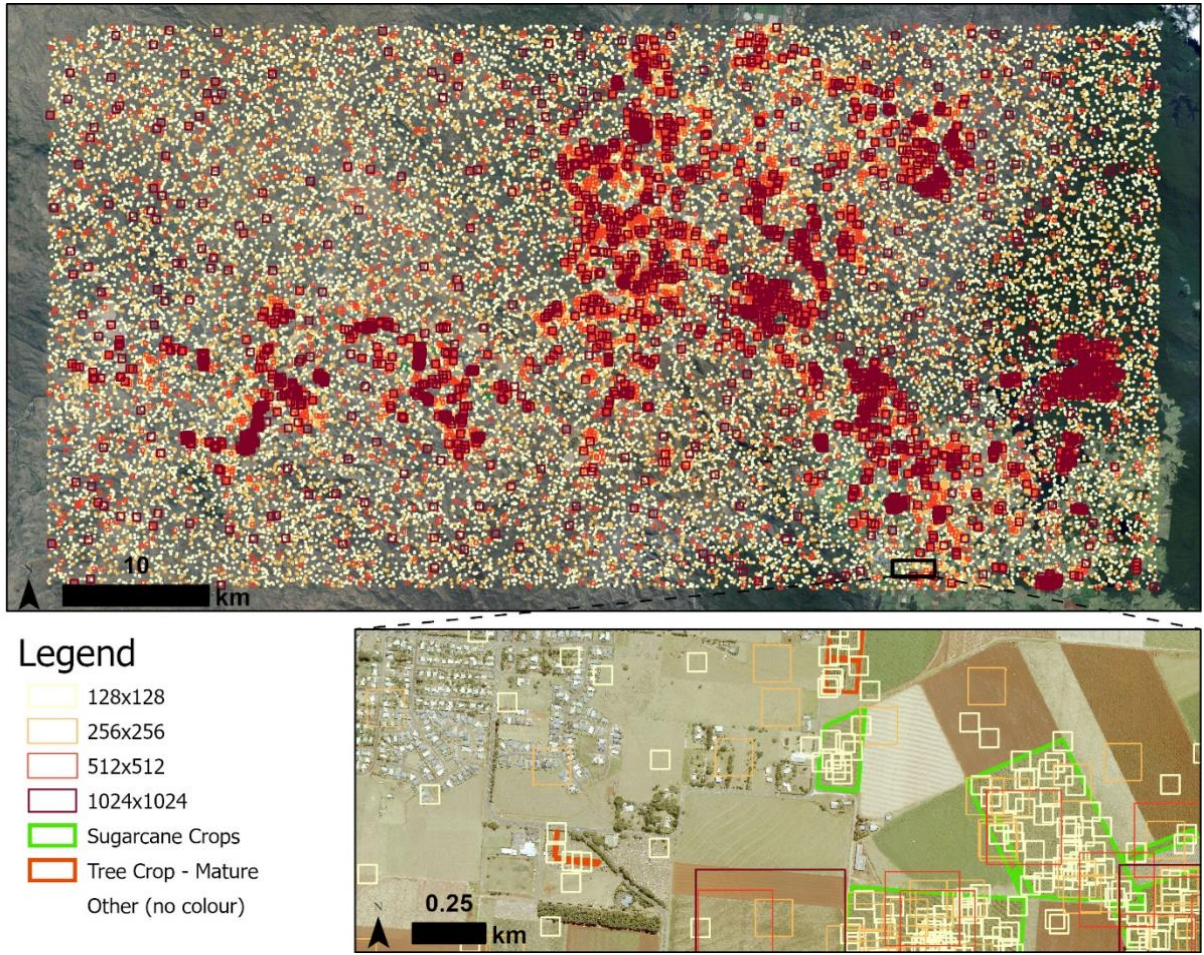

*Figure 5: Patch layout from the stratified random sampling strategy. Training classes show are derived from the manually derived training dataset.*

When training a neural network, the model weights were updated after each batch of training data was fed to the GPU. Smaller batch sizes will update the model weights more often than larger batch sizes. However, larger batch sizes have more data to inform the update of the model weights. To control the effect of the batch size on the results, the number of iterations per epoch remained consistent for each of the patch size tests (Table 5). This was deduced by determining an optimal batch size for the patch size of 1024 which was capable of fitting within the GPU memory (batch size of 16) and multiplying the result by 4, $4^2$ and $4^3$ to calculate the batch size value for 512, 256 and 128 respectively. This resulted in 150.5 batches of data for each epoch (rounded to 151).

*Table 5: Parameters for setting up the log number patches experiments. The number of iterations is calculated by dividing the number of patches by the batch size and rounded to the nearest whole number.*

| Size | Number of Patches | Batch Size | Batches per epoch |
|---|---|---|---|
| $1024^2$ | 2,408 | 16 | 151 |
| $512^2$ | 9,632 | 64 | 151 |
| $256^2$ | 38,528 | 256 | 151 |
| $128^2$ | 154,112 | 1,024 | 151 |

### 2.5.3 Data Augmentations

Aerial imagery can have poor calibration and varying imagery quality and resolution, particularly between capture dates because the same vendor, aircraft, camera and camera condition may not be used. As a result, spectral reflectance and spatial distortions can affect the appearance of features within the data. In addition, varying climatic conditions can also affect the spectral reflectance of features. To attempt to capture these variations, the training data can be augmented by flipping, rotating and changing the brightness of the image (Dosovitskiy et al., 2013; Wieland et al., 2019), which creates a more robust model for these image types and prevents overfitting of the data (Kattenborn et al., 2021).

The python package imgaug v0.4.0 ([Alexander Jung, 2020](#)) was used to apply random augmentations to the training data. The types of augmentation chosen were dependent on whether they were deemed useful for remote sensing applications. The augmentations selected were based on:

- altering the contrast and colourations (gamma, sigmoid, AllChannelsCLASHE, linear, multiply, allChannelsHistogramEqualization)
- adding noise to the image (salt and pepper, multiply element wise, additive gaussian, additive Poisson, multiply – different for each channel)
- and altering the geometry and scale of the image by zooming and stretching the image (affine, elastic transformation, vertical and horizontal flips)
- adding blur and artificial clouds/fog/smoke to mimic varying environmental and climatic conditions, different resolutions, capture angles and aircraft roll effects which are not always full corrected in provided imagery.

For each training image patch, contrast, noise and geometric distortions are applied at random and 50% of the time either blur or artificial clouds/fog is applied.

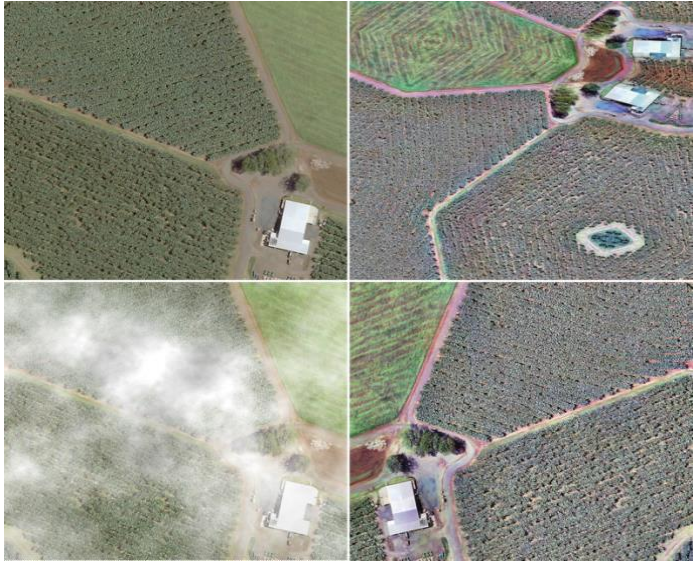

*Figure 6: Example of data augmentations over an area consisting of banana plantations including the original image (top left) and three augmentations versions. Note these examples do not represent the patch size used in this study but are for demonstration of the augmentations used for each patch.*

### 2.5.4 Data Scaling

Imagery from earth observation data can be supplied with a variety of pixel depths including integers, floating point numbers and a variety of sizes such as 8 bit or 32 bit and can include negative or positive only values. Without patch data scaling, models trained with, for example, pixel values between -1 and 1, will not transfer to imagery with values between 0 and 255. The imagery used in this project was supplied with 8-bit data type; however, the distribution of these data varied between the images (Figure 7). The data were scaled between 0 and 255 for each batch of patches.

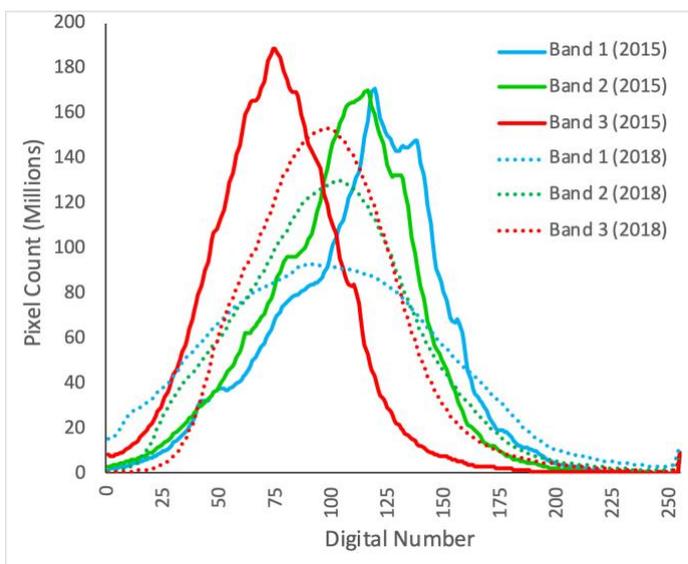

*Figure 7: Pixel value distribution for 2015 (solid) and 2018 (dashes) image bands.*

### 2.5.5 Multiple Pass Prediction

It has been found in previous studies (Sun et al., 2019) that the edges of each image chip have a lower accuracy than the centre region. To overcome this, a two-pass ensemble inference strategy was trialled. This was achieved by iteratively applying the model to the original image patch and averaging the resulting prediction from three rotated versions. The second pass of predictions was conducted offset by half a patch. The results from the two passes were combined using a weighted average based on distance, with pixels towards the centre of the patch given a higher weight than the pixels towards the edge.

### 2.5.6 Patch image and label generation

For all trials except for those for patch size and sampling, 22 830 patch extents were generated. The extents were used to extract training image chips. Corresponding labels were generated by converting the training polygon features to a raster representation covering the extent of the image chip. The information for each class was added to a separate band within the label file known as one hot encoding where 1 represents the presence of the class and 0 absence. As a result, the label rasters consisted of nine bands representing each of the classes in alphabetical order.

## 2.6 Training

The purpose of the training stage was to allow the model to learn how to identify land use classes. This was achieved by iterating the training images and labels to determine their relevant colour, texture and context attributes (Zhang et al., 2016). Each trial consisted of 20 iterations of the training data or epochs.

### 2.6.1 U-Net Architecture

The aim of the training was to produce a model to label every pixel in the image through semantic segmentation using a convolutional neural network. The structure of the CNN was based on the U-Net architecture (Ronneberger et al., 2015). It consists of two parts, an encoding stage which down-samples the resolution of the input images and a decoding stage which up-samples and restores the images to the original resolution.

At each level of the encoding stages, two convolution operations were applied and a 2×2 max pooling operation used to down-sample the input images. The first level consisted of the original satellite image and label patches where a specified number of filters were applied. For this study we used 32 initial filters. At each subsequent level of the encoding side of the U-Net, the number of filters was doubled and the resolution halved until reaching the bottom level where 512 filters were applied with a 16 times reduction in resolution and pixels. Figure 8 is a graphical representation of the U-Net architecture from Clark & McKechnie, 2020.

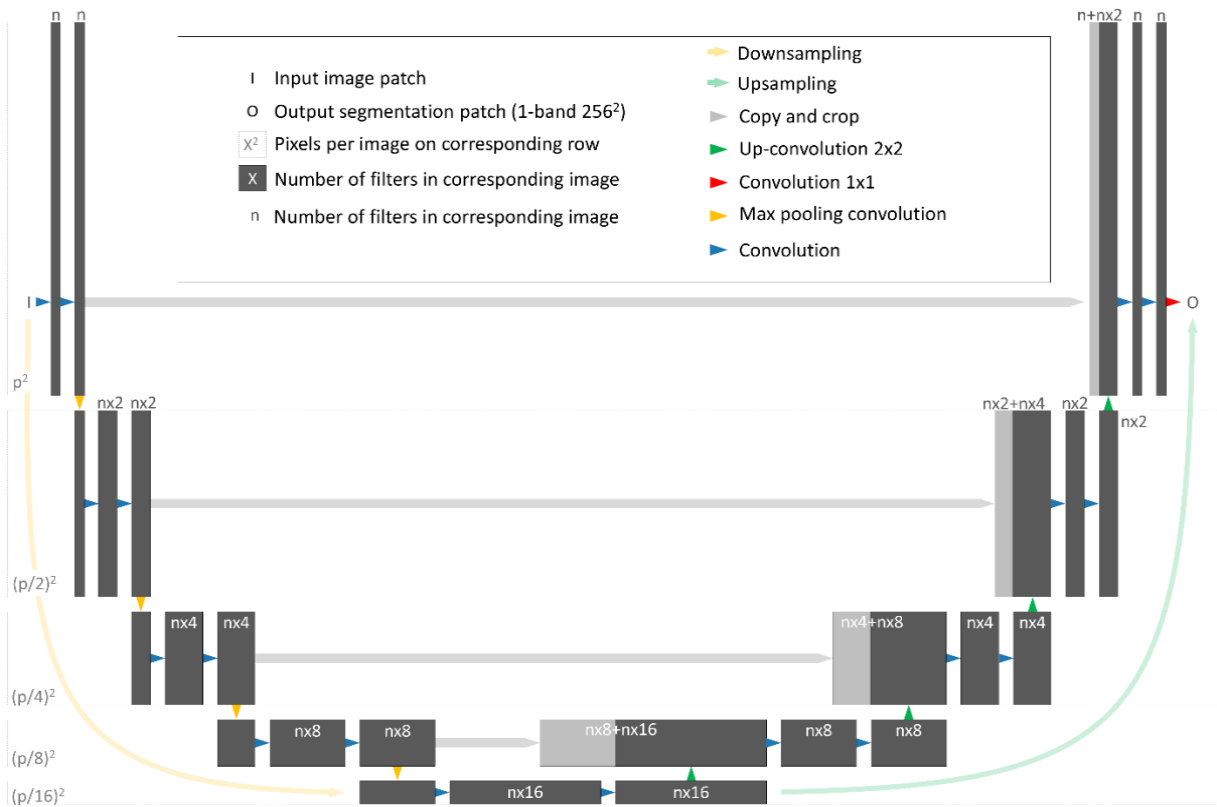

*Figure 8: The U-Net architecture (Clark & McKechnie, 2020; Ronneberger et al., 2015)*

## 2.7 Prediction

Classifications for each trial were produced for the 2018 training to assess how well a particular parameter learns the training data. The classifications for the fully trained models based on the optimised parameters were produced for 2015 and 2018. The output from the model prediction is raster with values between 0 and 1 for the nine classes represented in nine image bands. The prediction rasters are then flattened to a single band thematic map with the class containing the highest value considered the most probable feature within the image.

## 2.8 Accuracy Assessment

An independent accuracy assessment was conducted by randomly generating 10,000 points in an unbiased sampling approach. At each point, an observation was made for the 2018 and 2015 imagery and classified according to the project classes. Figure 9 shows the spread of data and classification according to the 2018 observation.

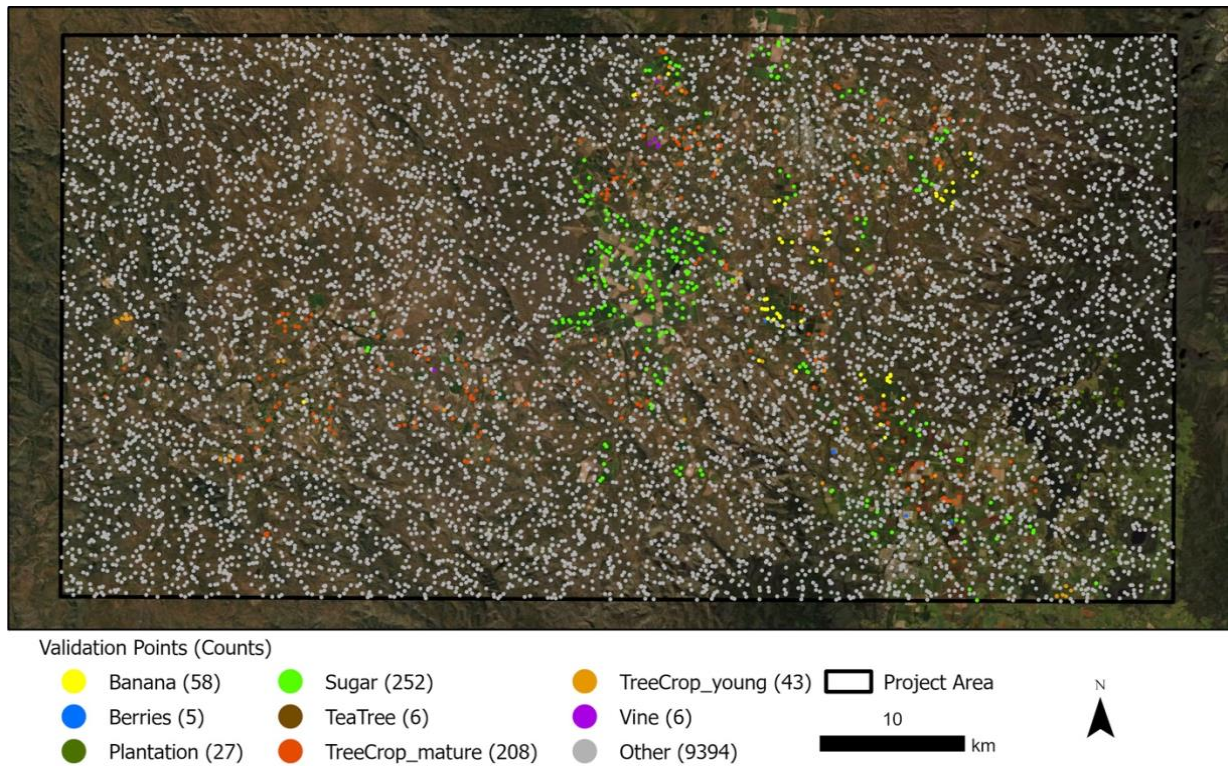

*Figure 9: Validation points used to assess the accuracy of the model classification coloured according to the 2018 observation value.*

Although 10,000 points were generated, the disproportionate area between classes resulted in 94% of the points being located within the 'Other' class. This means smaller classes such as berry crops, vineyards and tea tree plantations only have five or six points. However, as the trials were repeated five times, this resulted in a minimum of 25-point observations used to calculate the statistics for each class. Although it would be ideal to assess the accuracy using additional points, time and resource considerations limited this capacity.

The hand-crafted training data and each trial classification were compared against the validation points and accuracy was assessed by calculating the kappa and the user's accuracy (precision), producer's accuracy (recall) and F1-score metrics for each class.

2.9   Ranking the Trials

Generally, more complex models are likely to perform better but require a considerable amount of time and resources. To assist in the interpretation of the results, the models were ranked by considering the reliability utilising user's (recall) and producer's (precision) accuracy and model training time. To account for slight changes in computation time due to uncontrolled factors such as computing infrastructure load, the times were rounded up to the nearest 15 minutes. Time was only considered a factor if there was more than 15 minutes between the minimum and maximum.

For each test, the metrics were ranked from one to the total number of tests. Higher ranking represents higher accuracy or lower computation time. The results were scored by adding the user's and producer's accuracies, and twice the time ranks (resulting in the time having equal weighting as accuracy). Each test was assigned a final ranking based on this score.

## 2.10 Full training of top performing models

Using the top trial rankings, full training was conducted to assess the highest possible result for the project area by training on 2018 data and applying the prediction to the 2015 data. As with the above trials, this was repeated five times but trained for 100 epochs.

## 2.11 Computing infrastructure and software

The Queensland Department of Environment and Science owns and operates High Performance Computer (HPC) facilities. The HPC infrastructure consists of 2256 threads, 8.8TB of memory, eight Nvidia Tesla V100 GPUs and NVMe drives which were used to process the training data, train the CNN model and to create the model inference.

The processing of vector and image data used the Geospatial Data Abstraction Library (GDAL) version 3.1.0 (https://gdal.org/) and the deep-learning part of the project utilised TensorFlow 2.1.0 (Abadi et al., 2016). Image augmentations used the python library imgaug 0.4.0 (https://imgaug.readthedocs.io/en/latest/).

# 3   Results and Discussion

The aim of this project was to provide guidance on how to collect and process training data for use in deep-learning projects involving earth observation data. The following sections present and discuss the results and provide recommendations on how other projects may best undertake processing of the training data to produce the best possible results. Table 6 shows the results for all trials for the 2018 training image.

*Table 6: Trial results for the 2018 training image.*

| Trial | Parameter | Average Training Time (hours) | Kappa (95% CI) | Average F1 - Score (95% CI) | Average User's Accuracy (Precision) | Average Producer's Accuracy (Recall) | Ranking |
|---|---|---|---|---|---|---|---|
| **Human Derived** | Manual | >200 | 0.96 | 0.95 | 0.97 | 0.92 | - |
| **Batch Size** | 10 | 1.2 | 0.67 (0.62 - 0.71) | 0.68 (0.56 - 0.74) | 0.63 (0.5 - 0.71) | 0.84 (0.66 - 0.92) | 5 |
|  | 50 | 0.8 | 0.63 (0.62 - 0.65) | 0.62 (0.6 - 0.64) | 0.53 (0.51 - 0.55) | 0.91 (0.89 - 0.92) | 2 |
|  | 100 | 0.8 | 0.55 (0.48 - 0.62) | 0.56 (0.54 - 0.59) | 0.49 (0.46 - 0.53) | 0.85 (0.81 - 0.89) | 1 |
|  | 150 | 0.8 | 0.43 (0.31 - 0.49) | 0.51 (0.44 - 0.56) | 0.43 (0.38 - 0.49) | 0.83 (0.81 - 0.86) | 5 |
|  | 200 | 0.8 | 0.41 (0.33 - 0.49) | 0.45 (0.41 - 0.49) | 0.38 (0.34 - 0.41) | 0.82 (0.81 - 0.83) | 4 |
|  | 250 | 0.7 | 0.45 (0.33 - 0.51) | 0.48 (0.43 - 0.51) | 0.42 (0.37 - 0.44) | 0.79 (0.77 - 0.8) | 2 |
|  | 280 | 0.7 | 0.36 (0.19 - 0.46) | 0.41 (0.31 - 0.45) | 0.36 (0.29 - 0.4) | 0.75 (0.65 - 0.82) | 5 |
| **Patch Size (Systematic)** | 128 | 3.7 | 0.69 (0.65 - 0.77) | 0.49 (0.37 - 0.59) | 0.53 (0.35 - 0.66) | 0.51 (0.42 - 0.62) | 2 |
|  | 256 | 1.3 | 0.58 (0.4 - 0.72) | 0.48 (0.34 - 0.56) | 0.49 (0.34 - 0.61) | 0.55 (0.42 - 0.63) | 2 |
|  | 512 | 1.2 | 0.7 (0.62 - 0.74) | 0.49 (0.45 - 0.52) | 0.53 (0.46 - 0.61) | 0.53 (0.46 - 0.61) | 1 |
|  | 1024 | 1.3 | 0.66 (0.48 - 0.75) | 0.42 (0.41 - 0.45) | 0.43 (0.4 - 0.46) | 0.47 (0.42 - 0.52) | 4 |
| **Patch Size (stratified-random)** | 128 | 3.7 | 0.42 (0.38 - 0.49) | 0.46 (0.41 - 0.5) | 0.38 (0.35 - 0.43) | 0.86 (0.85 - 0.87) | 4 |
|  | 256 | 1.2 | 0.51 (0.47 - 0.58) | 0.51 (0.47 - 0.57) | 0.43 (0.4 - 0.49) | 0.86 (0.85 - 0.86) | 2 |
|  | 512 | 1.2 | 0.65 (0.59 - 0.71) | 0.62 (0.52 - 0.67) | 0.59 (0.48 - 0.68) | 0.77 (0.7 - 0.81) | 1 |
|  | 1024 | 1.3 | 0.53 (0.31 - 0.66) | 0.54 (0.38 - 0.62) | 0.5 (0.36 - 0.56) | 0.71 (0.52 - 0.82) | 3 |
| **Data Augmentation** | FALSE | 2.8 | 0.73 (0.7 - 0.77) | 0.7 (0.66 - 0.74) | 0.64 (0.58 - 0.68) | 0.87 (0.83 - 0.9) | 1 |
|  | TRUE | 8.9 | 0.49 (0.34 - 0.59) | 0.5 (0.48 - 0.53) | 0.43 (0.4 - 0.46) | 0.75 (0.72 - 0.77) | 2 |
| **Data Scaling** | FALSE | 2.1 | 0.69 (0.62 - 0.74) | 0.64 (0.56 - 0.69) | 0.59 (0.48 - 0.66) | 0.78 (0.72 - 0.81) | 2 |
|  | TRUE | 2.1 | 0.77 (0.7 - 0.79) | 0.74 (0.68 - 0.78) | 0.68 (0.6 - 0.74) | 0.88 (0.83 - 0.91) | 1 |
| **Multiple Pass Prediction** | Single | 8.9 | 0.49 (0.34 - 0.59) | 0.5 (0.48 - 0.53) | 0.43 (0.4 - 0.46) | 0.75 (0.72 - 0.77) | 2 |
|  | Multiple | 8.9 | 0.55 (0.4 - 0.65) | 0.52 (0.48 - 0.53) | 0.45 (0.41 - 0.48) | 0.75 (0.69 - 0.81) | 1 |

## 3.1 Training Data

The collection of training data took over 200 hours to manually hand digitise. This is a significant challenge for many semantic segmentation applications (Qin & Liu, 2022) and potentially negates any benefits of deep-learning approaches. The availability of the existing banana dataset (Clark & McKechnie, 2020) assisted greatly and it is recommended future studies leverage existing datasets where possible.

Table 7 lists the 2018 training data collected for this project. Approximately 94% of the project area consisted of the 'other' class while only 0.03% of the area contained berry crops. This class imbalance is very typical for projects classifying LULC using earth observation data.

*Table 7: Number of features, area and proportion of the project area for each class for 2018.*

| Name | Feature Count | Area (ha) | Area (%) |
|---|---|---|---|
| Banana Plantation | 243 | 1,860 | 0.62 |
| Berry Crops | 69 | 92 | 0.03 |
| Forestry Plantation | 118 | 981 | 0.33 |
| Sugarcane Crop | 515 | 7,621 | 2.54 |
| Tea Tree Plantation | 42 | 188 | 0.06 |
| Tree Crop - Mature | 2,289 | 6,249 | 2.09 |
| Tree Crop - Young | 280 | 988 | 0.33 |
| Vineyards | 33 | 146 | 0.05 |
| Other | 323 | 281,344 | 93.95 |
| **Total** | **3,912** | **299,471** | **100.00** |

The ability for a model to successfully train with high accuracy is reliant on the accuracy of the training data. If the training data is of poor quality, the model may not be able to determine the ideal weights for the model neurons to achieve the highest accuracy for classification. To a certain extent, CNN models may account for some level of error in the training data but may result in the model being penalised for achieving a higher accuracy than the data used to assess its performance (Burke et al., 2021).

The training data used in the project has its own inaccuracies as maintaining focus while hand digitising features over extended periods of time can be problematic (Van Coillie et al., 2014). As shown in Table 6 and Table 8, the human derived training data achieved an F1-Score of 0.95. Analysing these data to the class level revealed some individual classes such as the young tree crops class achieved an F1-Score of only 0.73 (Table 8) which limits the ability of the model to identify this class.

*Table 8: Per class accuracy of the human derived classifications for 2018 and 2015.*

| Class | 2018 | | | 2015 | | |
|---|---|---|---|---|---|---|
| | F1-Score | User's (Precision) | Producer's (Recall) | F1-Score | User's (Precision) | Producer's (Recall) |
| Banana Plantations | 1.0000 | 1.0000 | 1.0000 | 0.9908 | 0.9818 | 1.0000 |
| Berry Crops | 1.0000 | 1.0000 | 1.0000 | 0.8571 | 1.0000 | 0.7500 |
| Plantation Forestry | 0.8400 | 0.9545 | 0.7500 | 0.8780 | 1.0000 | 0.7826 |
| Sugarcane Crops | 0.9839 | 1.0000 | 0.9683 | 0.9204 | 0.9946 | 0.8565 |
| Tea Tree Plantation | 1.0000 | 1.0000 | 1.0000 | 1.0000 | 1.0000 | 1.0000 |
| Tree Crop - Mature | 0.9602 | 0.9361 | 0.9856 | 0.9403 | 0.9141 | 0.9679 |
| Tree Crop - Young | 0.7297 | 0.8710 | 0.6279 | 0.6897 | 0.8696 | 0.5714 |
| Vineyards | 1.0000 | 1.0000 | 1.0000 | 1.0000 | 1.0000 | 1.0000 |
| Other | 0.9981 | 0.9973 | 0.9989 | 0.9968 | 0.9950 | 0.9987 |
| **Total** | **0.9458** | **0.9732** | **0.9256** | **0.9192** | **0.9728** | **0.8808** |
| **Kappa** | **0.9617** | | | **0.9293** | | |

During the training data collection process, separating young and mature tree crops was extremely subjective and quite difficult, and was reflected within the accuracy assessment of training data, and as a result, in the accuracy of resulting models outlined below. As a comparison, the banana plantation class was based on a previous deep-learning model developed by Clark & McKechnie, 2020 resulting in an F1-Score of 1 for 2018 and 0.99 for 2015 (Table 8). The classes with smaller areas such as tea tree plantations and vineyards had a high accuracy as these features were easily identifiable.

The sugarcane crop class was complicated by several challenges. Firstly, the 2018 training image was captured in August, early in the sugarcane harvest season when cane was either fully mature or yet to be planted. The dates for the 2015 test image were captured as late as October when some of the fields harvested early in the season contained young sugarcane. The collection of the training data for 2015 was consistent with the 2018 training data and only mature sugarcane was included. However, for the accuracy assessment points, we decided to call the validation point sugarcane if the canopy of the crop was predominantly closed.

The sugarcane crop class was further complicated by the presence of maize crops in the southeast of the project area. Maize can look similar to sugarcane and was only separated during the training and validation data collection through local knowledge and identification of farm management practices only seen at the broad scale and not within a single image patch. Ideally, maize should be included as a separate class; however, due to time constraints, it remained as part of the 'other' class. There were also areas of abandoned sugarcane crops which were not in production but were still identified by the models.

### 3.2  Batch Size

The batch size results indicate a higher accuracy using smaller batch sizes (Table 6). During the training process, model weights are updated at the end of every batch which results in models with smaller batch sizes updating their weight more often likely resulting in more refinement of model weights and faster convergence. These results are consistent with Kandel & Castelli (2020) who also recommended using lower learning rates with small batch sizes although data scaling and augmentation may assist training when using larger batch sizes (Keskar et al., 2017).

Based on the results of batch size in combination with patch size results, it is recommended to increase patch size while decreasing batch size to achieve higher accuracy.

### 3.3  Patch Size and Sampling Strategy

A challenging factor within remote sensing applications is class imbalance from over or underrepresented classes in the training data (Maxwell et al., 2021). For the project area in this study, the 'other' class represented 94% of the area whereas the berry crop class only represented 0.03%. Systematic or random generation of training patches will have very few training samples for underrepresented classes, resulting in their poorer classification. In addition, different areas within class features mean features with smaller areas will be sampled less than larger features, resulting in their poorer classification.

Other studies have applied a class weighting (Daudt et al., 2019); however, this does not solve the problem of under-sampled classes and features. Classes with smaller area will still contain very few patches and remain underrepresented within the training data.

Results from the systematic grid sampling method (Table 6) did not indicate a significant increase in accuracy for each patch size. The stratified random sampling method showed the 512 x 512 pixel patch size produced higher accuracy for both the kappa statistic and F1-Score compared to other patch sizes. The kappa statistic indicated a slight reduction in accuracy compared to the grid sampling strategy (2018: -0.06); however, the F1-Score improved (2018: +0.13).

The stratified random sample approach is a workable solution to overcome the class imbalance issue. The overall results for the patch size and sampling strategies do not indicate a significant improvement in accuracy between the trials; however, the largest improvements were produced in the smaller classes particularly for larger patch sizes.

Figure 10 shows an example of this improvement for the vine class. The grid sampling strategy tests showed inferior performance with some models not able to classify the vine features. In contrast, the stratified random sample strategy was able to detect the vines in most cases with some degree of accuracy. Based on these results, the best performance achieved used a patch size of 512 x 512 pixels.

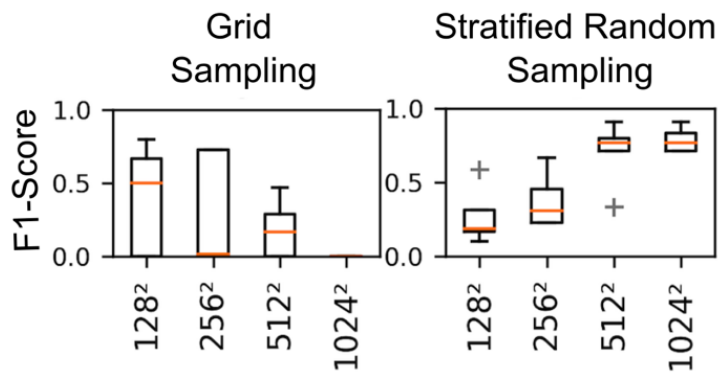

*Figure 10: Box and whisker plots showing grid-based and stratified random sampling strategies for the vineyard class F1-scores.*

## 3.4 Data Augmentation

The data augmentation trials (Table 6) decreased model accuracy (kappa: -0.24; F1-Score: -0.2) and increased training time. This result is expected as the random augmentations have avoided overfitting the model to the training data through presenting an altered version of the training data each time the data is loaded. As a result, we decided to assess the model on the 2015 test image (Table 9). We found when comparing a models trained with and without augmentations, there was a >0.24 increase in kappa and 0.19 increase in F1-Score. This demonstrates how to avoid overfitting the model to the training data to increase the transferability to unseen data.

*Table 9: Augmentation trial results for the 2015 test image.*

| Parameter | Average Training Time (hours) | Kappa (95% CI) | Average F1-Score (95% CI) | Average User's Accuracy (Precision) | Average Producer's Accuracy (Recall) | Ranking |
| --- | --- | --- | --- | --- | --- | --- |
| False | 2.8 | 0.12 (0.09 - 0.15) | 0.21 (0.18 - 0.25) | 0.22 (0.18 - 0.27) | 0.35 (0.31 - 0.39) | 2 |
| True | 8.9 | 0.36 (0.27 - 0.41) | 0.4 (0.38 - 0.41) | 0.37 (0.33 - 0.4) | 0.61 (0.57 - 0.65) | 1 |

Although applying augmentations increased the training time (x3), it created a more robust model which allowed for better transferability to other data.

The results from this trial indicate the importance of using a range of image augmentations to alter image perspective, colour and brightness. The pixel value variations within the 2018 training image and 2015 test images result from different camera configurations, post processing of the image tiles into a seamless mosaic and atmospheric and climatic conditions. These are all typical occurrences for earth observation data, particularly when using high spatial resolution data from aerial photography or satellites.

It is recommended any project attempting to ensure model transferability not only to a different time but also to a different sensor or geographic region implement data augmentations.

## 3.5 Data Scaling

Table 6 shows the kappa statistic improved by 0.08 and the F1-Score by 0.1 when patch data scaling was applied. Although only one type of scaling was tested for this project, other scaling and normalising options are possible such as scaling between 0 and 1. We used the range of 0 - 255 as to maintain compatibility with the imgaug library which recommends pixel values in this range. We recommend the implementation of data scaling particularly for projects using multiple imagery sensors.

## 3.6 Multiple Pass Prediction

Comparing single and multiple pass prediction methods (Table 6) shows the multiple pass method marginally improves prediction accuracy (kappa difference: +0.06; F1-Score difference: +0.02). However, the multiple pass classification is more aesthetic with the elimination of patch edge effects (Figure 11).

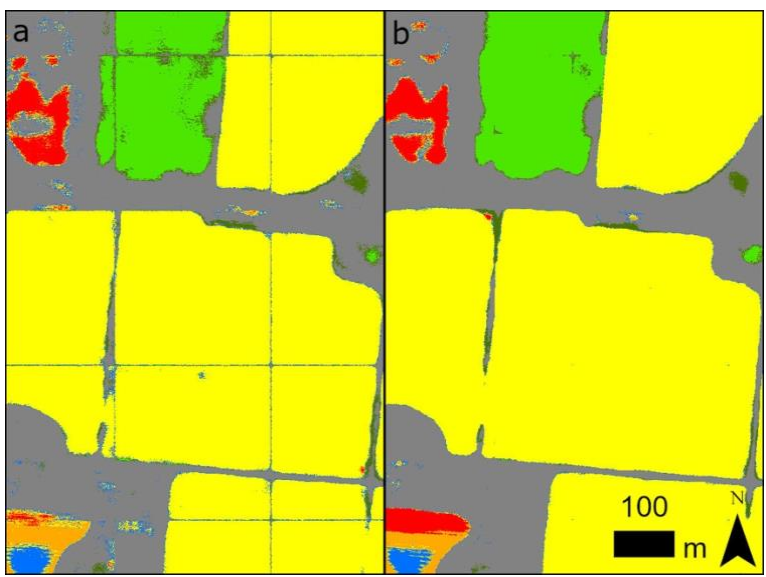

*Figure 11: An example of a single pass prediction (a) compared to a multiple pass prediction and augmentations (b).*

Although only a marginal improvement in accuracy resulted from the multiple pass method, this is recommended as it results in a smoother final classification.

## 3.7 Full Training

The objective of this study was to determine the optimal pre-processing steps to maximise the transferability of a model to a different sensor and different date. Based on result from the trials, five models were trained for 100 epochs using the parameters presented in Table 10.

*Table 10: Parameters used for the 100 epoch training trials.*

| Parameter | Value |
| --- | --- |
| Patch size | 512 |
| Sampling strategy | Stratified random sample (area) |
| Number of patches | 22,830 |
| Batch size | 20 |
| Data Augmentations | True |
| Data Scaling | True |
| Multiple Pass Prediction | True |

Training the models for 100 epochs resulted in the models achieving a kappa statistic of 0.9 (0.89-0.91) and 0.84 (0.82-0.87), a user accuracy of 0.8 (0.78-0.83) and 0.78 (0.76-0.8) and a producer accuracy of 0.98 (0.98-0.98) and 0.87 (0.85-0.9) for 2018 and 2015 respectively. As the models now contain the optimal pre-processing steps for the training data and are fully trained, we will no longer discuss the 2018 results.

*Table 11: Resulting accuracy measures for the 100 epoch training trials.*

| Image | Kappa (95% CI) | Average F1-Score (95% CI) | Average User's (95% CI) (Precision) | Average Producer's (95% CI) (Recall) |
| --- | --- | --- | --- | --- |
| 2018 | 0.9 (0.89 - 0.91) | 0.87 (0.86 - 0.89) | 0.8 (0.78 - 0.83) | 0.98 (0.98 - 0.98) |
| 2015 | 0.84 (0.82 - 0.87) | 0.81 (0.79 - 0.84) | 0.78 (0.76 - 0.8) | 0.87 (0.85 - 0.9) |

Figure 12 shows the confusion matrix for 2015. The model performs well at finding all land use features; however, there is some confusion between the tree crop classes (mature and young) and between the other class and the sugarcane, tree crops (mature and young), vineyards and tea tree classes.

*Figure 12: Full training trial number 5 2015 classification confusion matrix*

The per class results for all five trials (Figure 13) showed most classes achieved an accuracy >70% except for the vineyards class. The main confusion was with areas of fallow. The 2018 training data contained many noticeably young vineyards where vines were barely evident in the image. This resulted in some vineyard features resembling areas of ploughed fallow. Only 0.5% of the project area contained the vineyard class (Table 7) which was the second smallest class in area. Berries were the smallest class; however, most berries in the project area were contained within a greenhouse which makes these features easily identifiable within the imagery.

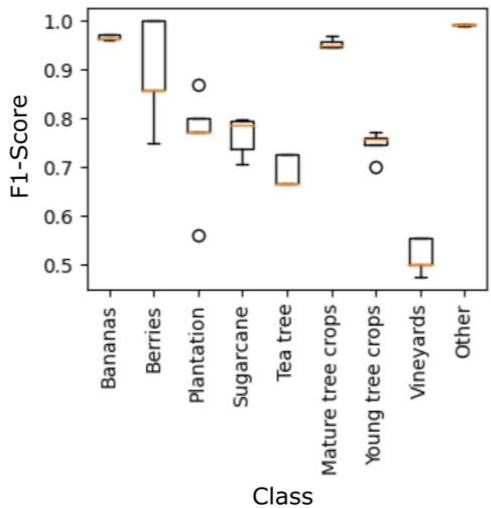

*Figure 13: Box and whisker plots for F1-Scores for each class from the five full trained models.*

Figure 14 shows the human derived (Figure 14a) and 2015 (Figure 14b) output classifications. At this scale, confusion between sugarcane crops and other land uses is evident.

Figure 15 shows the 2015 and 2018 imagery, hand digitised 2015 validation and 2018 training data and resulting classifications for the top performing of the five fully trained models. This model achieved a kappa of 0.87 and 0.84 and an F1-score of 0.84 and 0.77 for 2018 and 2015 respectively.

The results showed some of the inaccuracies within the training and validation data. The 2015 manually classified data for the young tree crops example in Figure 15 showed areas of missed tree crops which were identified by the model classification consistent with the findings of other studies using noisy data (Qin & Liu, 2022). The young tree crop example also identified an area of confusion between young and mature tree crops in the 2018 training data which led to confusion with the model results (Figure 12).

Figure 15 also demonstrates an area of emerging sugarcane crops in 2015 which was not included in the manual classification as it was deemed the canopy had not fully closed as discussed in section 0. In addition, it is also evident in Figure 15 the misclassification of sugarcane crops with the 'other' class as demonstrated in Figure 12.

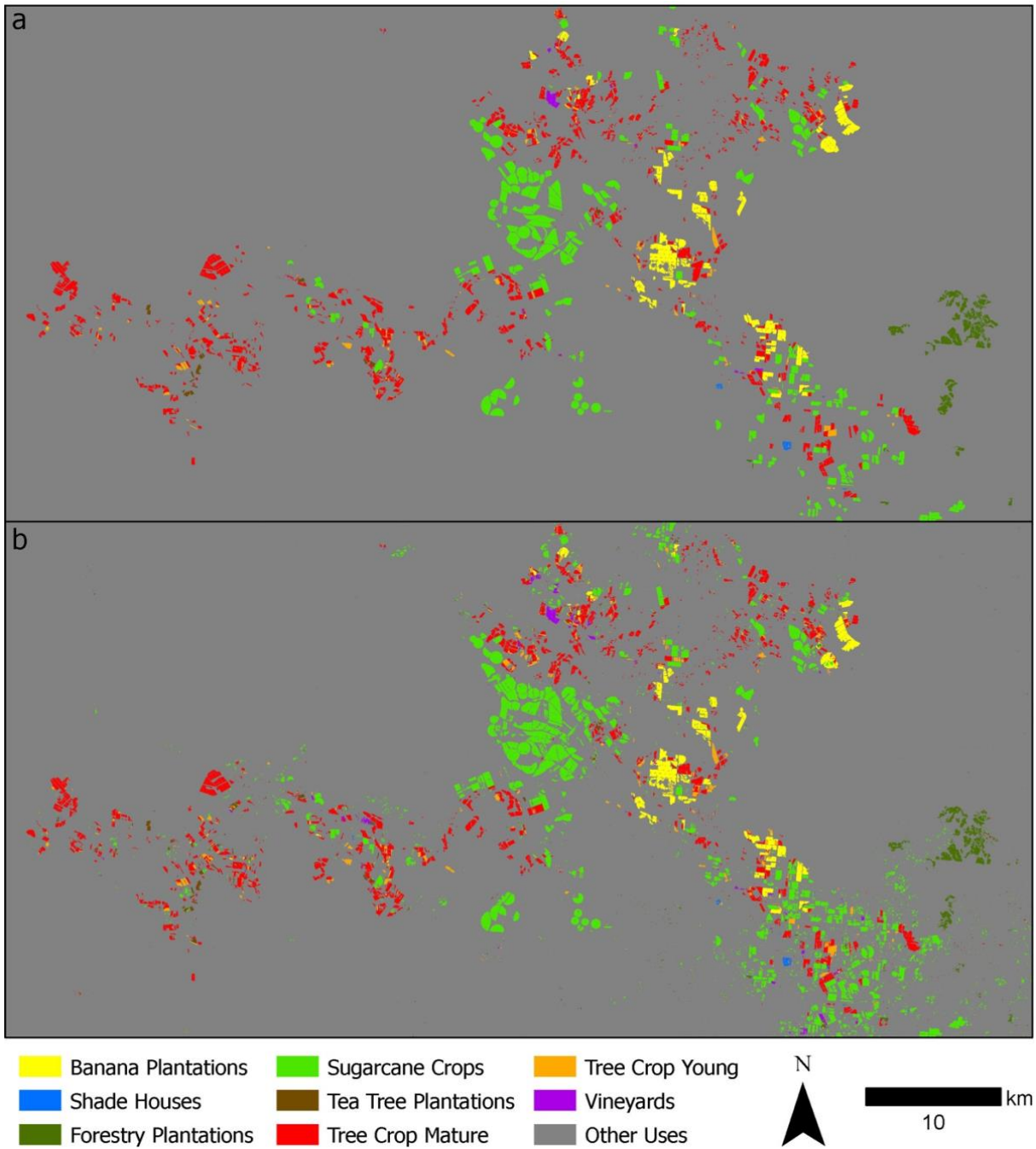

*Figure 14: Comparison for the project area for human classification (a) and prediction result for 2015 (b).*

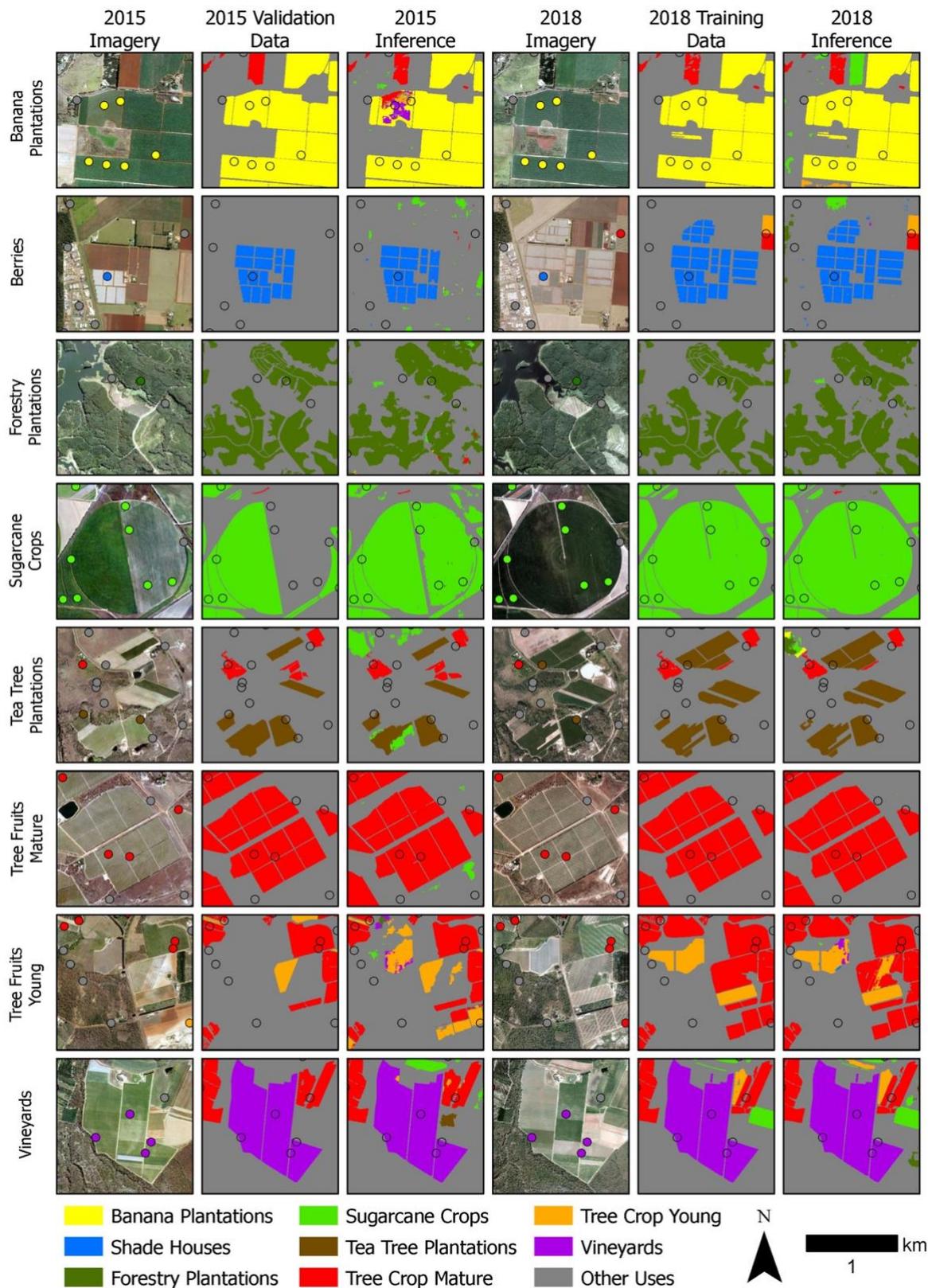

*Figure 15: 2015 and 2018 imagery, manual classification and prediction result for a model trained for 100 epochs. The 2018 manual classification was used to derive the training data for the model. The circles indicate the location of the accuracy assessment points with the colour in the imagery columns representing the class according to the legend.*

## 4 Limitations and future research

This study restricted the analysis to three band 50 cm imagery. These results may not be applicable for training a model for earth observation applications for different spatial and spectral resolutions. Convolutional neural networks are suited to higher resolution data (<1m) (Liu et al., 2019; Wurm et al., 2019) although some success has been achieved using earth-i (Flood et al., 2019) sentinel data (Stoian et al., 2019). It is expected presented here such as sampling strategy, data augmentations and multiple pass prediction will still be applicable at higher resolutions however resolution specific aspects such as patch size (and as a result batch size) will need to be re-evaluated.

Additional research could investigate different approaches to producing the output classifications. In this study we have used a multiple-pass strategy involving overlapping patches and a weighted mean to counteract patch edge effects when producing the output classification, however there are alternatives such as trimming of edge pixels and ensuring the patches overlap by the same number of pixels as presented in Flood et al. (2019).

There are several outstanding questions which need to be address in future research. First, we did not classify all LULC features within the project area due to time restrictions. Future work should analyse the effect on model accuracy with additional LULC classes. It would also be interesting to compare the training time and model accuracy when a model is created for each class separately in contrast to having one model for all classes.

In this study, we compared generating the output classification using a single pass strategy to a multiple pass with augmentations. Although the multiple pass method produced a more accurate and aesthetic classification, we did not compare the additional time it took to generate the classification. We did not examine if averaging the patch classification with three augmented versions has any advantage over one or two augmented version. Producing outputs efficiently is imperative for timely delivery of broad scale LULC classifications.

The production of LULC classification often rely on additional ancillary data and existing experience of the skilled professional. The integration of additional ancillary information such as, for example, climate, elevation and soil information may assist in the prediction of LULC features, specifically over broad areas.

## 5 Conclusion

The objective of this study was to better understand how to prepare earth-observation data for training of a multi-class deep-learning model. Firstly, we recommend the use of existing datasets where available. The collection of training data for this project took hundreds of hours even with using existing datasets and prior knowledge of the area. Although freely available datasets usually do not match the spatial resolution used in projects implementing CNNs, the training of deep-learning models can tolerate a certain level of noise within the data and, in some cases, the trained model may have a higher level of accuracy compared to the original training data.

The most substantial improvements in the transferability of the model from the 2018 training image to the 2015 test image resulted from image augmentations and scaling of the data. Data augmentation and scaling are imperative to avoid overfitting the model to the training data, model generalisation and transferability, and therefore are recommended.

Compared to the grid sampling approach, the stratified random sampling approach for generating image patches substantially increased inaccuracy for small classes in our imbalanced training dataset. This approach, although not improving overall model accuracy metrics, substantially improved accuracy in detecting classes which represent only a small proportion of the landscape. For projects with class imbalances, it is recommended this sampling strategy be implemented.

When producing image patches, we recommend generating larger patch sizes and training with a lower batch number rather than smaller patches with larger batches.

Applying the model to imagery using different perspectives through patch rotations and applying a second pass over the prediction image with a half patch offset improves the output accuracy and creates a more aesthetically pleasing classification.

**Authors' contributions**

Formal analysis, A.C.; investigation, A.C.; data curation, A.C.; writing—original draft preparation, A.C.; writing—review and editing, A.C., S.P., P.S.; visualization, A.C.; supervision, S.P. and P.S.; project administration, A.C.

**Acknowledgements**

The Authors would like to acknowledge the support from the Australian Government Research Training Program Scholarship and the Queensland Government for supplying the data used in this paper and for use of the High Performance Computing Infrastructure.

# 6 References

Abadi, M., Agarwal, A., Barham, P., Brevdo, E., Chen, Z., Citro, C., Corrado, G. S., Davis, A., Dean, J., Devin, M., Ghemawat, S., Goodfellow, I., Harp, A., Irving, G., Isard, M., Jia, Y., Jozefowicz, R., Kaiser, L., Kudlur, M., … Zheng, X. (2016). TensorFlow: Large-Scale Machine Learning on Heterogeneous Distributed Systems. *ArXiv:1603.04467*. http://arxiv.org/abs/1603.04467

Bai, X., Sharma, R. C., Tateishi, R., Kondoh, A., Wuliangha, B., & Tana, G. (2017). A Detailed and High-Resolution Land Use and Land Cover Change Analysis over the Past 16 Years in the Horqin Sandy Land, Inner Mongolia. *Mathematical Problems in Engineering*, *2017*, 1–13. https://doi.org/10.1155/2017/1316505

Ball, J. E., Anderson, D. T., & Chan, C. S. (2017). Comprehensive Survey of Deep Learning in Remote Sensing: Theories, Tools, and Challenges for the Community. *Journal of Applied Remote Sensing*, *11*(04), 1. https://doi.org/10.1117/1.JRS.11.042609

Blaschke, T., Hay, G. J., Kelly, M., Lang, S., Hofmann, P., Addink, E., Queiroz Feitosa, R., van der Meer, F., van der Werff, H., van Coillie, F., & Tiede, D. (2014). Geographic Object-Based Image Analysis – Towards a new paradigm. *Isprs Journal of Photogrammetry and Remote Sensing*, *87*(100), 180–191. PMC. https://doi.org/10.1016/j.isprsjprs.2013.09.014

Burke, M., Driscoll, A., Lobell, D. B., & Ermon, S. (2021). Using Satellite Imagery to Understand and Promote Sustainable Development. *Science*, *371*(6535), eabe8628. https://doi.org/10.1126/science.abe8628

Caye Daudt, R., Le Saux, B., Boulch, A., & Gousseau, Y. (2019). Multitask learning for large-scale semantic change detection. *Computer Vision and Image Understanding*, *187*, 102783. https://doi.org/10.1016/j.cviu.2019.07.003

Clark, A., & McKechnie, J. (2020). Detecting Banana Plantations in the Wet Tropics, Australia, Using Aerial Photography and U-Net. *Applied Sciences*, *10*(6), 2017. https://doi.org/10.3390/app10062017

Deng, L. (2014). A Tutorial Survey of Architectures, Algorithms, and Applications for Deep Learning. *APSIPA Transactions on Signal and Information Processing*, *3*. https://doi.org/10.1017/atsip.2013.9

Dosovitskiy, A., Springenberg, J. T., & Brox, T. (2013). Unsupervised feature learning by augmenting single images. *ArXiv:1312.5242 [Cs]*. http://arxiv.org/abs/1312.5242

DSITI. (2017). *Land Use Summary 1999–2015 for the Atherton Tablelands* (p. 26). Department of Science, Information Technology and Innovation, Queensland Government. https://publications.qld.gov.au/dataset/land-use-summary-1999-2015/resource/d97bee40-5694-424a-9085-c7e4892475b8

Flood, N., Watson, F., & Collett, L. (2019). Using a U-Net Convolutional Neural Network to Map Woody Vegetation Extent from High Resolution Satellite Imagery Across Queensland, Australia. *International Journal of Applied Earth Observation and Geoinformation*, *82*, 101897. https://doi.org/10.1016/j.jag.2019.101897

Hey, A. J. G. (Ed.). (2009). *The fourth paradigm: Data-intensive scientific discovery*. Microsoft Research.

Hoeser, T., & Kuenzer, C. (2020). Object Detection and Image Segmentation with Deep Learning on Earth Observation Data: A Review-Part I: Evolution and Recent Trends. *Remote Sensing*, *12*(10), 1667. https://doi.org/10.3390/rs12101667

Jensen, J. R. ; (2007). *Remote Sensing of the Environment* (2nd ed.). Prentice Hall.

Kandel, I., & Castelli, M. (2020). The effect of batch size on the generalizability of the convolutional neural networks on a histopathology dataset. *ICT Express*, *6*(4), 312–315. https://doi.org/10.1016/j.icte.2020.04.010

Kattenborn, T., Leitloff, J., Schiefer, F., & Hinz, S. (2021). Review on Convolutional Neural Networks (CNN) in vegetation remote sensing. *ISPRS Journal of Photogrammetry and Remote Sensing*, *173*, 24–49. https://doi.org/10.1016/j.isprsjprs.2020.12.010

Keskar, N. S., Mudigere, D., Nocedal, J., Smelyanskiy, M., & Tang, P. T. P. (2017). On Large-Batch Training for Deep Learning: Generalization Gap and Sharp Minima. *5th International Conference on Learning Representations*. http://arxiv.org/abs/1609.04836

Lillesand, T. M., Kiefer, R. W., & Chipman, J. W. (2015). *Remote sensing and image interpretation* (Seventh edition). John Wiley & Sons, Inc.

Liu, P. (2015). A survey of remote-sensing big data. *Frontiers in Environmental Science*, *3*, 45. https://doi.org/10.3389/fenvs.2015.00045

Liu, S., Qi, Z., Li, X., & Yeh, A. (2019). Integration of Convolutional Neural Networks and Object-Based Post-Classification Refinement for Land Use and Land Cover Mapping with Optical and SAR Data. *Remote Sensing*, *11*(6), 690. https://doi.org/10.3390/rs11060690

Ma, L., Liu, Y., Zhang, X., Ye, Y., Yin, G., & Johnson, B. A. (2019). Deep Learning in Remote Sensing Applications: A Meta-Analysis and Review. *ISPRS Journal of Photogrammetry and Remote Sensing*, *152*, 166–177. https://doi.org/10.1016/j.isprsjprs.2019.04.015

Ma, Y., Wu, H., Wang, L., Huang, B., Ranjan, R., Zomaya, A., & Jie, W. (2015). Remote sensing big data computing: Challenges and opportunities. *Future Generation Computer Systems*, *51*, 47–60. https://doi.org/10.1016/j.future.2014.10.029

Maxwell, A. E., Warner, T. A., & Guillén, L. A. (2021). Accuracy Assessment in Convolutional Neural Network-Based Deep Learning Remote Sensing Studies—Part 2:

Recommendations and Best Practices. *Remote Sensing*, *13*(13), 2591. https://doi.org/10.3390/rs13132591

Neupane, B., Horanont, T., & Hung, N. D. (2019). Deep learning based banana plant detection and counting using high-resolution red-green-blue (RGB) images collected from unmanned aerial vehicle (UAV). *PLOS ONE*, *14*(10), e0223906. https://doi.org/10.1371/journal.pone.0223906

Pandey, P. C., Koutsias, N., Petropoulos, G. P., Srivastava, P. K., & Dor, E. B. (2021). Land Use/Land Cover in View of Earth Observation: Data Sources, Input Dimensions, and Classifiers—a Review of the State of the Art. *Geocarto International*, *36*(9), 957–988. https://doi.org/10.1080/10106049.2019.1629647

Qin, R., & Liu, T. (2022). A Review of Landcover Classification with Very-High Resolution Remotely Sensed Optical Images—Analysis Unit, Model Scalability and Transferability. *Remote Sensing*, *14*(3), 646. https://doi.org/10.3390/rs14030646

Ronneberger, O., Fischer, P., & Brox, T. (2015). U-Net: Convolutional Networks for Biomedical Image Segmentation. *ArXiv:1505.04597 [Cs]*. http://arxiv.org/abs/1505.04597

Stoian, A., Poulain, V., Inglada, J., Poughon, V., & Derksen, D. (2019). Land Cover Maps Production with High Resolution Satellite Image Time Series and Convolutional Neural Networks: Adaptations and Limits for Operational Systems. *Remote Sensing*, *11*(17), 1986. https://doi.org/10.3390/rs11171986

Sun, Y., Tian, Y., & Xu, Y. (2019). Problems of Encoder-Decoder Frameworks for High-Resolution Remote Sensing Image Segmentation: Structural Stereotype and Insufficient Learning. *Neurocomputing*, *330*, 297–304. https://doi.org/10.1016/j.neucom.2018.11.051

Van Coillie, F. M. B., Gardin, S., Anseel, F., Duyck, W., Verbeke, L. P. C., & De Wulf, R. R. (2014). Variability of operator performance in remote-sensing image interpretation: The importance of human and external factors. *International Journal of Remote Sensing*, *35*(2), 754–778. https://doi.org/10.1080/01431161.2013.873152

Wieland, M., Li, Y., & Martinis, S. (2019). Multi-Sensor Cloud and Cloud Shadow Segmentation with a Convolutional Neural Network. *Remote Sensing of Environment*, *230*, 111203. https://doi.org/10.1016/j.rse.2019.05.022

Wurm, M., Stark, T., Zhu, X. X., Weigand, M., & Taubenböck, H. (2019). Semantic Segmentation of Slums in Satellite Images Using Transfer Learning on Fully Convolutional Neural Networks. *ISPRS Journal of Photogrammetry and Remote Sensing*, *150*, 59–69. https://doi.org/10.1016/j.isprsjprs.2019.02.006

Zang, N., Cao, Y., Wang, Y., Huang, B., Zhang, L., & Mathiopoulos, P. T. (2021). Land-Use Mapping for High-Spatial Resolution Remote Sensing Image Via Deep Learning: A Review. *IEEE Journal of Selected Topics in Applied Earth Observations and Remote Sensing*, *14*, 5372–5391. https://doi.org/10.1109/JSTARS.2021.3078631

Zhang, L., Zhang, L., & Du, B. (2016). Deep Learning for Remote Sensing Data: A Technical Tutorial on the State of the Art. *IEEE Geoscience and Remote Sensing Magazine*, *4*(2), 22–40. https://doi.org/10.1109/MGRS.2016.2540798